# Open-Vocabulary Semantic Segmentation Network Integrating Object-Level Label and Scene-Level Semantic Features for Multimodal Remote Sensing Images

Jinkun Dai[a,b], Yuanxin Ye[a,b], Peng Tang[a,b], Tengfeng Tang[a,b], Xianping Ma[a,b], Jing Xiao[c], Mi Wang[d]

[a]Faculty of Geosciences and Engineering, Southwest Jiaotong University, Chengdu 611756, China,

[b]State-Province Joint Engineering Laboratory of Spatial Information Technology for High-Speed Railway Safety,

Southwest Jiaotong University, Chengdu 611756, China

[c]School of Artificial Intelligence, Wuhan University, Wuhan 430072, China

[d]State Key Laboratory of Information Engineering in Surveying, Mapping and Remote Sensing, Wuhan University, Wuhan 430072, China

**Abstract**：

Semantic segmentation of multi-modal remote sensing imagery plays a pivotal role in land use/land cover (LULC) mapping, environmental monitoring, and precision earth observation. Current multi-modal approaches mainly focus on integrating complementary visual modalities (e.g., optical and synthetic aperture radar (SAR) imagery), yet neglect the incorporating of non-visual textual data – a rich source of knowledge that can bridge semantic gaps between visual patterns and real-world concepts. To address this limitation, we propose TSMNet, a text supervised multi-modal open vocabulary semantic segmentation network that synergistically integrates textual supervision with visual representation for open-vocabulary semantic segmentation. Unlike conventional multi-modal segmentation frameworks, TSMNet introduces a dual-branch text encoder to extract both scene-level semantic and object-level label information from various textual data, enabling dynamic cross-modal fusion. These text-derived features dynamically interact with visual embeddings through the proposed text-guided visual semantic fusion module, enabling domain-aware feature refinement and human-interpretable decision-making. Moreover, integrating text opens pathways for open-vocabulary semantic segmentation, enabling systems to recognize and classify unseen categories through natural language descriptions, thereby overcoming the rigid constraints of predefined class taxonomies. To verify our method, we innovatively construct two new multi-modal datasets, and carry out extensive experiments to make a comprehensive comparison between the proposed method and other state-of-the-art (SOTA) semantic segmentation models. Results demonstrate that TSMNet achieves superior segmentation accuracy while exhibiting robust generalization capabilities across diverse geographical and sensor-specific scenarios. This work establishes a new paradigm for explainable remote sensing analysis, demonstrating that textual knowledge integration significantly enhances model generalizability. The source code will be available at https://github.com/yeyuanxin110/TSMNet

**KEYWORDS**：Remote Sensing, Multi-modal, Open Vocabulary, Semantic Segmentation, Vision-language Model.

## 1. INTRODUCTION

Semantic segmentation is an advanced remote sensing technique that aims to perform pixel-wise classification of each pixel in the image, achieving pixel-level image segmentation (Xiao et al., 2025). It has become indispensable for critical Earth observation tasks such as environmental monitoring (Li et al., 2020), urban planning (Liu et al., 2018; Wu et al., 2022), disaster response (Sundaresan et al., 2025) and land use classification (Chen et al., 2017; Piramanayagam et al., 2018).

The proliferation of multi-source remote sensing datasets (e.g., optical, Light Detection and Ranging (LiDAR), Synthetic Aperture Radar (SAR)) driven by advances in sensor technology has revolutionized Earth observation paradigms (Zhu et al., 2017). However, the deluge of data not only drives methodological innovation but also imposes unprecedented challenges on interpretation methods (Xiong et al., 2025). In the past decade, deep learning has become a powerful approach to addressing complex tasks (LeCun et al., 2015; Luo et al., 2022); however, most studies still concentrate on single-modal semantic segmentation (Long et al., 2015; Ronneberger et al., 2015; Chen et al; 2018; Wang et al., 2020).

Single-modal approaches suffer from limited discriminative power when differentiating spectrally similar classes (e.g., asphalt vs. shadow). To address this, some studies introduce additional modes to overcome the performance bottleneck. For example, the combination of optical image and SAR image can effectively reduce the influence of bad weather and speckle noise on semantic segmentation models. The existing multi-modal semantic segmentation achieves by using complementary visual modes but ignores the potential of non-visual modes such as text data. Textual data offers a transformative opportunity: it embeds real-world knowledge (e.g., geographic context, material properties) that can provide critical contextual priors for model robustness (Li et al., 2023; Li et al., 2025). Beyond that, the integrated text opens the way for semantic segmentation of open vocabulary (Wang et al., 2024a; He et al., 2023), which enables the system to identify and classify invisible categories through natural language description, thus overcoming the strict restrictions of predefined category classification (Kawano et al., 2024). Different from the traditional semantic segmentation method (Pan et al., 2025), which is limited to a fixed set of tags, open vocabulary semantic segmentation allows a wider range of concepts, making it more flexible and suitable for new scenarios in practical applications (Cheng et al., 2022).

Recently, Visual Language Models (VLMs) has attracted great attention because of its excellent open-vocabulary object recognition ability (Zhang et al., 2023a; Jose et al., 2024; Zhang et al., 2025). This great success motivates us to explore its adaptability to multi-modal semantic segmentation tasks. VLMs have demonstrated excellent feature representation ability through cross-modal contrastive learning (Valerie et al., 2023). However, this representation lacks pixel-level granularity, making it challenging to directly apply to dense prediction tasks. Inspired by the work of natural language processing, a series of methods are proposed in the field of vision, aiming at effectively adapting VLMs to the task of open-vocabulary semantic segmentation (OVSS) (Wang et al., 2024c; Lin et al., 2024). The existing main method is prompt learning, and the adapter is used to enhance the feature representation of learning. In the realm of land cover mapping, eliminating dependence on fixed label sets opens the door for novel map generation, in which the category information used can be directly defined by users (Zhang et al., 2023b; Cao et al., 2025). Using simple language to define classes allows individuals, especially those without remote sensing expertise, to easily create personalized maps tailored to their specific needs (Zhang et al., 2024; Xu et al., 2024).

Inspired by the potential of remote sensing VLMs (Wang et al., 2024b), we propose a text-supervised multimodal open vocabulary semantic segmentation network. Our framework aligns image-language features into a unified semantic space via contrastive learning, while enabling natural language interaction through hierarchical visual representation learning that leverages multi-layer high-level semantics. Specifically, TSMNet consists of the following components: multi-modal visual encoder for optical and SAR images, text encoder based on CLIP (Radford et al., 2021), scene-level semantic and object-level label fusion module of visual language. The multi-modal visual encoder of optical and SAR images adopts a pseudo-Siamese feature extraction module, and extracts multi-level semantic information of optical and SAR images respectively through vision transformer (ViT) (Dosovitskiy et al., 2021), and constructs a multi-modal remote sensing image feature fusion network to fully integrate the detailed features of multi-modal images. For language processing, TSMNet uses the Bidirectional Encoder Representations from Transformers (BERT) (Devlin et al., 2019), which is a universal language Transformer, to design a multi-layer image language fusion module, including scene-level semantic and object-level label features. The process begins by generating two distinct types of texts using a predefined prompt template. Text and image features are aligned via contrastive learning. Subsequently, the aligned text features are integrated with multi-modal image features through a text-guided visual semantic fusion module, ensuring rich contextual representation. TSMNet has gained strong representational capabilities of remote sensing images, enable deep exploration of the invariant features of images, and shows good generalization ability in semantic segmentation. The main contributions of TSMNet are summarized as follows:

1) In this paper, a TSMNet is developed, which innovatively integrates the fine-grained features of image and text modes, and realizes accurate semantic segmentation in open vocabulary scenes through multimodal feature interaction mechanism.

2) We design a dual-branch image and text fusion module (DITF), which integrates the image features with the scene-level semantic and object-level label features of the text by optimizing the text embedding, effectively integrates the heterogeneous graphic features, enhances the dependence within and between patterns, and thus enriches the semantic information.

3) To evaluate the model's generalization and practical value, we construct two visual language semantic segmentation datasets, which fill the key gap in the current multi-modal semantic segmentation dataset of integrated visual language. One of them is the semantic segmentation dataset of optical and SAR remote sensing images from Gaofen (GF) satellites, and the images are described manually. The dataset covers representative areas of diverse terrain, and each region is equipped with a high-precision ground truth label. On the other hand, we have meticulously annotated each group of images in an existing multi-modal remote sensing image dataset with detailed textual descriptions.

## 2. RELATED WORKS

### 2.1 Multi-modal semantic segmentation

Due to the limitations of imaging conditions, the information obtained from a single source image (e.g., optical image) has its constraints and cannot accurately describe the true state of the scene. It is susceptible to weather interference or insufficient structural details in complex terrains (Han et al., 2025). Therefore, it is necessary to introduce multi-source data to address the limitations of single-source data by capturing and fusing complementary information from different modalities. Driven by advancements in Earth observation technology, multi-modal remote sensing data such as optical, multispectral, hyperspectral and SAR images have become increasingly available. Optical images provide rich spectral and texture details but perform poorly in adverse weather conditions, while SAR data can penetrate clouds and capture structural information but suffer from speckle noise and low interpretability. Thus, fusing optical and SAR images can help improve overall segmentation performance (Ma et al., 2024).

Currently, how to extract and fuse features from various cross-modal images has become a research hotspot. As two widely used data sources in the remote sensing field, the fusion of optical and SAR data has attracted significant attention from researchers. The tremendous progress in deep learning has driven remarkable developments in the field of multi-modal semantic segmentation (Jiang et al., 2024). Many scholars have focused on studying multi-modal fusion methods. The simplest approach is to stack optical and SAR images into four channels as a whole input to a multi-modal semantic segmentation network, such as RseUNet-a (Diakogiannis et al., 2020). Early fusion methods relied on simple strategies such as feature concatenation and weighted averaging. Methods like PSCNN (Kuo et al., 2022) and MRSDC (Yang et al., 2020) concatenate features along the channel dimension, while FuseNet (Hazirbas et al., 2016) and vFuseNet (Audebert et al., 2017) use addition to fuse cross-modal features. However, simply stacking multi-modal data often fails to effectively capture cross-modal semantic relationships and is sensitive to noise. To address this, attention-based networks have been introduced to improve feature fusion. Li et al. (2022b) introduces a cross-modal attention module (MCANet), dynamically aggregating features of optical and SAR data by computing a channel affinity matrix, while CroFuseNet (Wu et al., 2023) adopts multi-scale segmentation branches to fuse hierarchical features through adaptive weighting. Additionally, CMGFNet (Hamidreza et al., 2022) proposes a gated fusion module that adaptively learns discriminative features and removes irrelevant information for effective fusion. However, existing frameworks often use global pooling layers to compute attention weights, which reduces spatial resolution and may lose fine-grained details crucial for segmentation.

Recently, the Transformer architecture has been widely adopted in multi-modal fusion for its strong contextual modeling and cross-modal integration capabilities (Ma et al., 2025). For example, CMFNet (Fan et al., 2023) introduces a multi-scale cross-modal transformation method for feature fusion. CEN (Li et al., 2021) proposes a parameter-free channel exchange network to adaptively share information between modalities. TokenFusion (Wang et al., 2022) further enhances fusion by dynamically detecting and replacing uninformative tokens with aggregated cross-modal features. Additionally, SwinFusion (Ma et al., 2022) first uses convolutional layers to extract shallow features and then employs the SA-based Swin Transformer to generate deep features. This indicates that Transformers can also serve as backbones for multi-modal fusion tasks.

While multi-modal data enriches remote sensing image interpretation with complementary information, it also introduces challenges. Different imaging mechanisms capture diverse and complex object features, and significant cross-modal semantic gaps hinder alignment in a shared feature space for stable fusion. Moreover, the current multimodal network lacks the assistance of text information, and it is difficult to achieve accurate semantic segmentation in complex geographical environment. Therefore, this paper constructs a TSMNet, which innovatively integrates the fine-grained features of visual and text modes, and realizes accurate semantic segmentation in open vocabulary scenes through multimodal feature interaction mechanism.

**2.2 Open-Vocabulary semantic segmentation**

OVSS aims to identify unseen category objects by leveraging semantic associations between vision and text (Zhou et al., 2022). Early works such as SPNet (Xian et al., 2019) and GroupViT (Xu et al., 2022) pioneered methods involving joint vision-text embedding spaces. SPNet projects pixel-level visual features into a semantic space aligned with fixed word embeddings, while GroupViT groups image regions under text supervision. However, due to the lack of unified multimodal pretraining, these methods suffer from misalignment issues, resulting in suboptimal semantic segmentation performance, yet also making the unified perception paradigm of OVSS highly anticipated.

The advent of VLMs like CLIP has revolutionized the field of OVSS (Ghiasi et al., 2022). At present, mainstream methods usually identify regions based on the similarity between image features and category text embedding. For example, LSeg (Li et al., 2022a) aligns CLIP's text embeddings with pixel-level features to achieve open-vocabulary classification. ZegFormer (Ding et al., 2022) employs a two-stage framework: a mask proposal network generates category-agnostic segmentation regions, and then CLIP classifies these regions using text embeddings. Leveraging CLIP's powerful generalization ability, CLIPseg and Freeseg follow prompt learning approaches. Based on this adapter-based method, the Side Adapter Network (SAN) introduces a side adapter network, which guide CLIP's deeper layers for proposal classification. Through an iterative cross-modal attention mechanism, SAN optimizes noisy text embeddings, achieving pixel-level semantic guidance without relying on external mask generators.

The Semantic-Assisted Calibration Network (SCAN) combines generalized semantic priors and context transfer strategies to enhance segmentation performance. In contrast, the Simple Encoder-Decoder Network (SED) emphasizes often-neglected local spatial details by using a CNN-based CLIP backbone to build an efficient OVSS network. Focusing on efficiency, the Global Knowledge Calibration Network (GKC) performs fast open-vocabulary segmentation by distilling knowledge from pretrained CLIP. However, these methods are primarily designed for natural images and do not adequately address the distinct characteristics of remote sensing imagery. To overcome this limitation and leverage the unique aspects of such data, we introduce an OVSS framework tailored specifically for remote sensing images.

## 3. METHODOLOGY

CLIP has shown great ability in open vocabulary classification. Nonetheless, a significant divergence exists between its image-level pretraining knowledge and the demands of pixel-level semantic segmentation, presenting a considerable challenge to bridge this domain gap. Fine-tuning CLIP directly on the downstream segmentation dataset will inevitably damage the ability of open vocabulary recognition. Therefore, our goal is to explore a method to achieve pixel-level alignment of image and text features, while preserving the alignment of scene features and text descriptions of images to enhance the model's comprehension of scene-level semantic and object-level label text information within remote sensing images. As depicted in Fig. 2, we propose a text-supervised multimodal open vocabulary semantic segmentation network. Firstly, we introduce an overview of the proposed TSMNet in Section 3.1. Then, in 3.2, we discuss how to better integrate the detailed features of multi-modal images. Subsequently, Section 3.3 details the object-level image-text feature fusion, and Section 3.4 presents the scene-level semantic fusion. Finally, the loss function and inference process are elaborated in Sections 3.5 and 3.6, respectively.

**3.1 Overall Network Architecture**

As illustrated in Fig. 1, the proposed network processes a set of input images $Input \in R^{H \times W \times 3}$, an $N$-vocabulary list and a set of text descriptions, where $H, W$ denote image dimensions and $N$ refers to the number of category names. The whole network processes data in three sequential stages, namely, the multi-modal image feature fusion stage, the object-level label feature fusion stage and scene-level semantic feature fusion stage.

To inherit the abundant pre-training knowledge from CLIP, we fuse the multi-scale pixel-level features from multi-modal images, then align fused image features with the corresponding text features of N vocabulary at pixel level and achieve alignment between visual and language representations by using context-aware prompting strategy. Finally, the fused visual features are integrated with the corresponding text embeddings to produce the semantic segmentation results of N vocabulary.

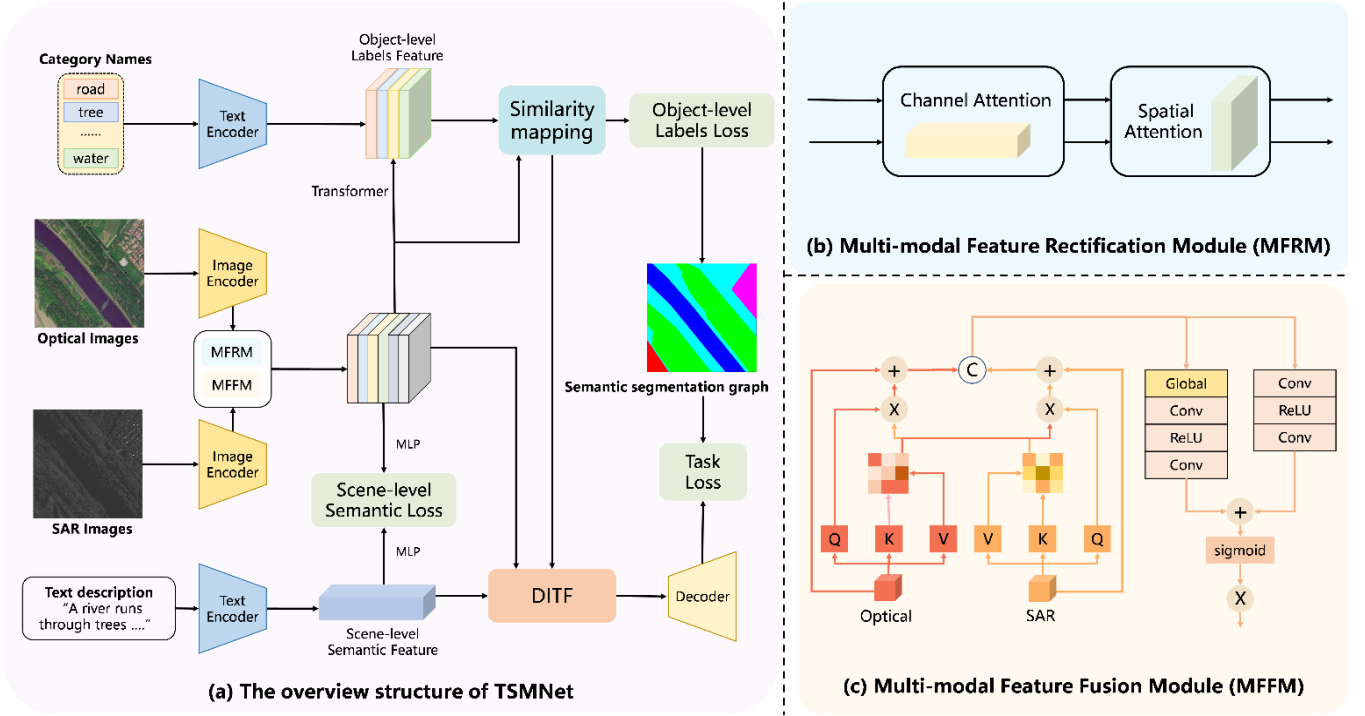

Fig. 1. The overall structure of TSMNet is shown in (a). TSMNet realizes the multi-modal open vocabulary semantic segmentation based on optical, SAR images and text data. The overall structure of MFRM is shown in (b). It dynamically adjusts the channel and spatial weights between multi-modal features to enhance the complementary information. The overall structure of MFFM is shown in (c). It introduces cross-modal attention mechanism for deep semantic interaction to realize multi-modal feature aggregation.

## 3.2 Multi-modal Image Feature Fusion Network

We propose a multi-modal image feature fusion network, which combines feature rectification and cross-modal interaction to address the challenges of fusing optical and SAR image features. Multi-modal data (such as optical and SAR) show significant heterogeneity, which makes it difficult for traditional linear weighted fusion methods to fully leverage the complementary information. Given the inherent complexity of multi-modal data, we adopt a ViT as our backbone feature extractor, and take the output of its four stages as multi-scale image features, which together with global image representation and local spatial features constitute image features. To achieve robust cross-modal integration, we devised a multi-modal feature rectification module (MFRM) and a multi-modal feature fusion module (MFFM) to fuse the six-layer features of the image respectively. The formulas are shown as follows:

$$x = ViT(X), y = ViT(Y), \tag{1}$$

$$\tilde{x}, \tilde{y} = MFRM(x, y), \tag{2}$$

$$z = MFFM(\tilde{x}, \tilde{y}), \tag{3}$$

where $X$ and $Y$ represent optical and SAR images respectively, $x$ and $y$ represent corresponding feature maps, $\tilde{x}$ and $\tilde{y}$ represent feature maps after multi-modal feature correction, and $z$ represents multi-modal fusion feature maps.

MFRM dynamically adjusts the channel and spatial weights between multi-modal features to enhance the complementary information. This hierarchical correction strategy effectively alleviates the difference of heterogeneous features between modes, and realizes the fine-grained alignment of channels and spatial dimensions, thus providing a highly consistent and discriminatory feature representation for subsequent cross-modal fusion. MFFM introduces cross-modal attention mechanism to realize deep semantic interaction. Multi-level fusion is implemented on multiple feature layers with different scales to preserve local details and global semantics. Multi-level feature aggregation is achieved through cross-modal attention and double-branch interaction. This module effectively solves the semantic conflict caused by direct fusion of heterogeneous features, and significantly improves the feature representation ability and task adaptability. Multi-stage fusion realizes fine information integration through hierarchical processing of multi-scale features. Firstly, in the feature extraction stage, dual-backbone network extracts multi-level features from both optical and SAR images. Then, in the stage of hierarchical correction and fusion, MFRM is used to eliminate modal deviation in turn, and MFFM is used for cross-modal interaction on multiple scales. Finally, in the output aggregation stage, the multi-level fusion features are integrated into a multi-scale feature pyramid. This strategy dynamically assigns weights and promotes cross-modal interaction, balancing local detail enhancement and global semantic consistency, thus improving the model's robustness in complex scenes and the generalization ability in multi-task scenes.

## 3.3 Object-level Label Feature Fusion of Image and Text

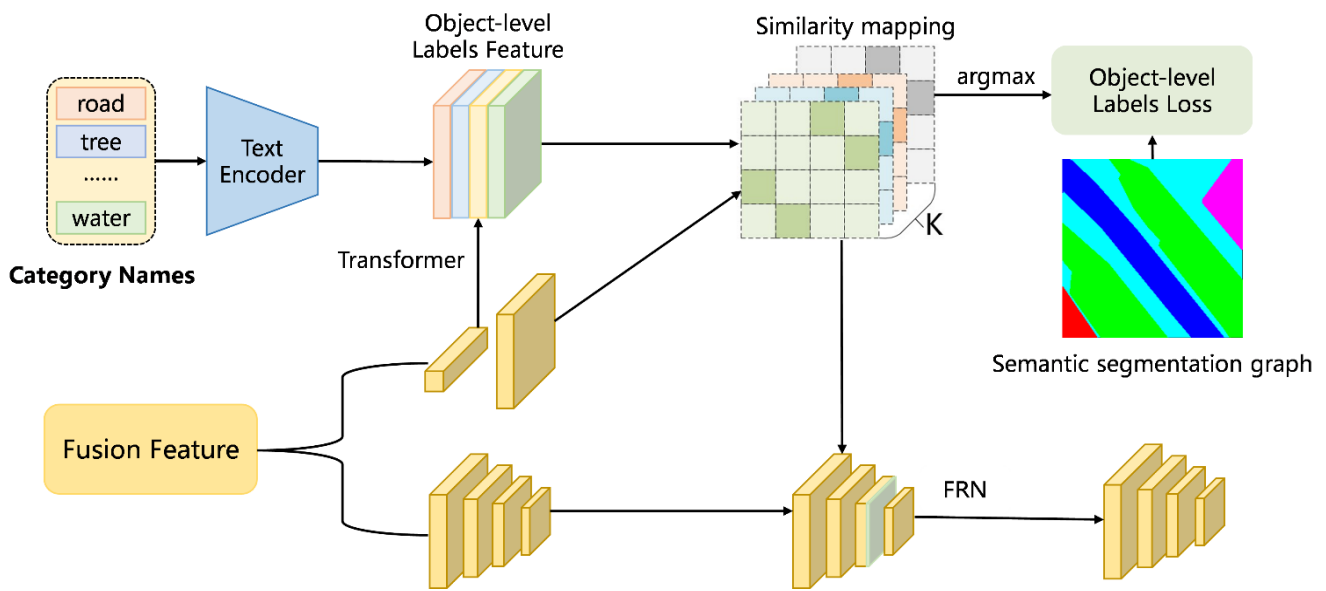

Fig. 2.  Object-level Label Feature Fusion of Image and Text.

As illustrated in Fig. 2, a fusion method of object-level image and text features is proposed, which uses the prior knowledge of language from CLIP pre-training model. CLIP uses two encoders: an image encoder and a text encoder. Visual and textual representations are aligned in a shared embedding space through contrastive learning on large-scale image-text pairs. It supports cross-modal understanding and zero-shot learning, providing a powerful foundational model for multimodal tasks. To transfer CLIP's prior knowledge to downstream semantic segmentation tasks, this paper adopts text prompts constructed based on templates. We construct text prompts by substituting the [CLS] placeholder in the template "a photo of a [CLS]" with K category names, and then encode it by the text encoder of CLIP.

Previous studies show that narrowing the gap between visual and language domains can greatly enhance the performance on subsequent tasks. Therefore, we attempt to adopt a context-aware approach to enhance text features, rather than relying solely on traditional predefined templates.

Consequently, the text encoder receives input as follows:

$$[p, e_k], 1 \leq k \leq K, \quad (4)$$

where $p$ is the learnable text context, and $e_k$ is the name embedding of the $k$-th category. Contextual features are composed of global image representations and local spatial features.

Including the description of visual context, can make the text more accurate. For example, "a dog on the grass" is more accurate than "a dog". We typically use the cross-attention mechanism in the decoder to model vision-language interaction. The context-aware strategy adopted in this paper is to processes the text features generated by the text encoder, that is, post-model prompt. We use template hints to generate text features, which are directly used in the query of transformer.

$$v = Transformer(t, [\bar{z}, z]), \quad (5)$$

where $t$ is the text feature and $z$ is the language-compatible multimodal fusion image feature.

This approach prompts text features to locate the most pertinent visual cues and subsequently refines these text features via the remaining connections:

$$t \leftarrow t + \gamma v, \quad (6)$$

where $\gamma$ is a trainable parameter that adjusts the size of the residual. It starts with a tiny value (like 0.0001) to preserve the initial linguistic knowledge from the text features as much as possible.

Next, we generate a pixel-text score map by computing the similarity between the language features $z$ and the text features $t$:

$$s = \hat{z}\hat{t}^T, \quad (7)$$

where $\hat{z}$ and $\hat{t}$ represent the $\ell_2$ normalized versions of $z$ and $t$, respectively, along the channel dimension.

The score map quantitatively represents the matching correspondence between pixel-level visual patterns and textual semantics. First, the score map can be regarded as a segmentation result, and we can use it to assist in calculating the segmentation loss. Second, we connect the score map to the final feature map to incorporate linguistic prior knowledge. Input it into a neck module based on Feature Pyramid Network (FPN) to fuse and enhance the features of different levels, so as to better support the tasks of target detection and segmentation.

### 3.4 Scene-level Semantic Feature Fusion of Image and Text

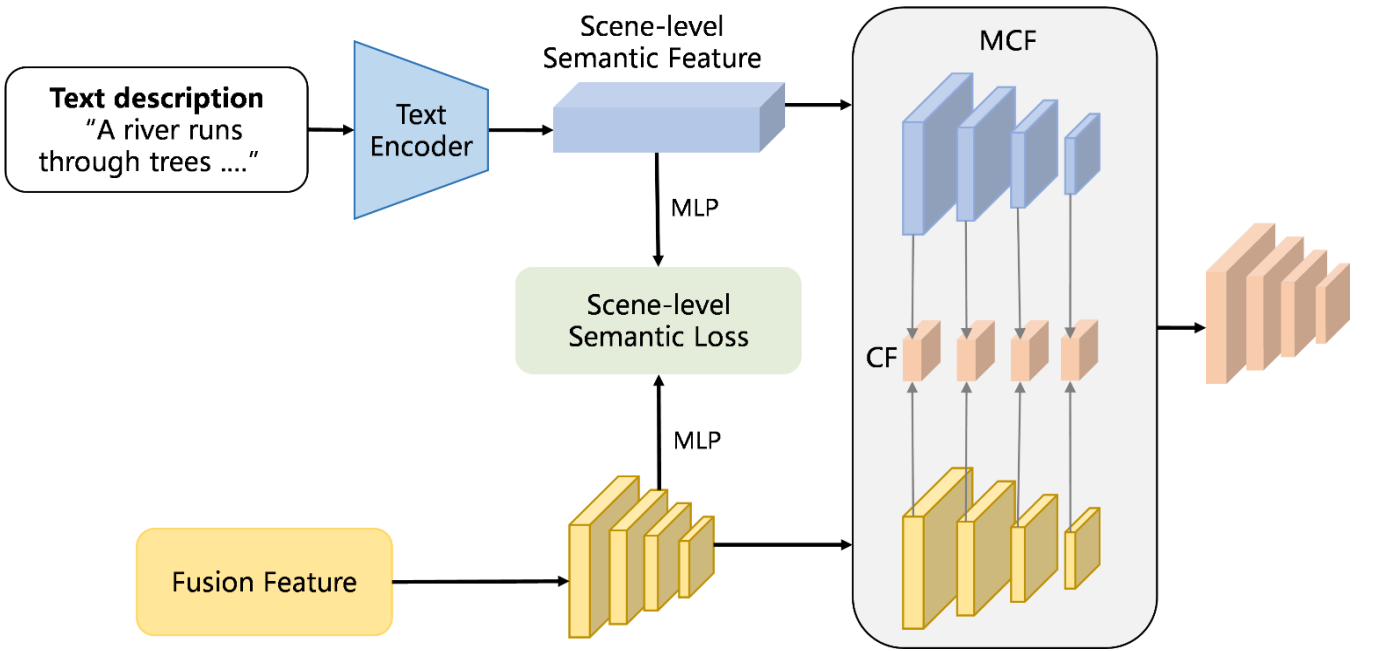

Fig. 3.  Scene-level Semantic Feature Fusion of Image and Text.

Compared with local category information, people often read global information from images first. Because there is a big gap between image and text description, cross-modal image-text feature alignment and fusion are the key factors in visual language representation learning. In order to solve this problem, we developed an global alignment and fusion module for image and text features, which consists of two main components, namely, image-text alignment module and image-text fusion module, as shown in Fig. 3.

1) Image-text alignment module: We adopt the strategy of " alignment first and then fusion " to deal with heterogeneous image-text features. We use the contrastive loss for image-text to bridge the gap between the two types of features. This step helps to establish an initial relationship between the image and text features, which in turn encourages their fusion during subsequent processing.

2) Image-text fusion module: Text features including climate information and geographical object features can be used as global priors for cross-modal feature fusion. Therefore, this paper proposes an image-text fusion module based on cross-modal attention mechanism, aiming at effectively combining image and text features and improving the feature representation ability in multi-modal tasks. The module firstly encodes the global text features and dynamically generates multi-scale text features to match the image features at different levels. Then, the cross-modal attention mechanism is used to fuse image features and text features layer by layer to capture the semantic association between them. Specifically, for each level of image features, the module interacts with the corresponding scale of text features through the cross-modal attention mechanism to generate a fused feature representation. Finally, the module outputs multi-scale fusion features to provide rich cross-modal information for downstream tasks. Experiments show that the module can significantly improve the performance of joint representation of images and texts, and provide an effective feature fusion solution for multimodal tasks.

**3.5 Loss Function**

In the training process of TSMNet, three main losses are designed and calculated: object-level label loss, scene-level semantic loss and semantic segmentation task loss. Object-level tag loss is used to guide the optimization of image and tag text information, scene-level semantic loss is used to guide the optimization of image and text description information, and semantic segmentation task loss ensures that the segmentation mask generated by the model is consistent with the target area. Specifically, we use cross-entropy loss for the object-level label loss, which is represented as follows:

$$\mathcal{L}_{Object} = CrossEntropyLoss = -\sum_{i=1}^{C} y_i \log(p_i), \qquad (8)$$

With:

$$p_i = \frac{\exp(z_i)}{\sum_{j=1}^{C} \exp(z_j)}, \qquad (9)$$

Where $C$ is the total number of classes, $p_i$ is the probability of class $i$ predicted by the model, and $z_i$ is the original output of the model to class $i$.

Scene-level semantic loss typically employs the InfoNCE (Noise Contrastive Estimation) loss, which is widely utilized in contrastive learning frameworks. The mathematical formulation of this loss function can be expressed as follows:

$$\mathcal{L}_{Scene} = -\log \frac{\exp(sim(z_i, z_t)/\tau)}{\sum_{j=1}^{N} \exp(sim(z_i, z_{t_j})/\tau)}, \tag{10}$$

where $z_i$ and $z_t$ are the embedding of image and text respectively, $sim(,)$ is the similarity function (such as cosine similarity), $\tau$ is the temperature parameter, and $N$ is the number of negative samples.

At the same time, we use cross-entropy loss ($\mathcal{L}_{ce}$) and dice loss ($\mathcal{L}_{dice}$) for the semantic segmentation task loss. The total loss of TSMNet is the sum of four types of losses, which can be expressed as follows:

$$\mathcal{L}_{total} = \mathcal{L}_{Object} + \mathcal{L}_{Scene} + \mathcal{L}_{ce} + \mathcal{L}_{dice}, \tag{11}$$

### 3.6 Inference

As shown in fig. 1, the network can generate the similarity mapping between category text and image while generating the segmentation map. In the reasoning stage, firstly, the input text categories are encoded by a text encoder, and the target embedding is obtained. Then, the cosine similarity between the semantic features of each pixel position of the target embedding and fusion features is calculated, so that the similarity mapping of each target text is generated. Finally, the final segmentation mask is obtained by selecting the label corresponding to the maximum of all similarity values at each pixel position. Compared with the traditional semantic segmentation method, this network has obvious advantages: it can not only predict fixed categories of goals, but also support open vocabulary setting. This means that users can use any English word or text fragment as the target category, which greatly enhances the flexibility and adaptability of the model. This ability of semantic segmentation of open vocabulary enables models to interact in real time by calculating cosine similarity of coded text prompts. This allows models to rapidly generate multiple similarity maps or segmentation masks. Therefore, the network can not only handle the semantic segmentation task of known categories, but also realize more flexible and dynamic segmentation in the open vocabulary scene, which provides a wider possibility for practical application.

# 4. EXPERIMENTS AND RESULTS

## 4.1. Data Description

In the field of deep learning, large-scale and high-quality data sets are indispensable for improving model performance and generalization ability. However, in the field of semantic segmentation of multimodal remote sensing images, there is still a lack of publicly available image text data sets. In order to solve this gap, we created two sets of data sets, one is SWJTU-Vision-Language dataset, and the other is YESeg-OPT-SAR dataset. These data sets cover the typical features of remote sensing, ensuring strong representativeness and applicability.

Table 1: Sample size of two multi-modal vision-language datasets (256 × 256 pixels).

| Dataset | All | Train | Test |
| --- | --- | --- | --- |
| SWJTU-Vision-Language | 2712 | 800 | 1912 |
| YESeg-OPT-SAR | 2231 | 800 | 1431 |

1) SWJTU-Vision-Language dataset: This dataset is a multi-modal and high-resolution remote sensing image dataset, which is specially designed for semantic segmentation tasks driven by deep learning. By integrating optical and SAR imagery from the same area, this dataset constructs a joint semantic segmentation dataset. The dataset consists of 2,712 pairs of multi-modal images with a resolution of 3 meters, covering four major cities including Beijing, Jinan, Shanghai and Tianjin and their surrounding areas. Each pair of images is manually labeled with detailed text descriptions, which not only provide semantic information of image content, but also help the model to better understand the relationship between different features. In order to further improve the practicability of the dataset, all images are labeled at pixel level, and each pixel is accurately divided into seven different categories. This fine labeling method makes the dataset very suitable for training and evaluating deep learning models, especially in complex land use segmentation tasks. Optical images provide rich color and texture information, while SAR images can penetrate clouds and vegetation to provide structural information of the ground. The combination of the two makes the data set maintain high segmentation accuracy even under cloudy or night conditions. The selection of Beijing, Jinan, Shanghai and Tianjin makes the dataset have extensive geographical coverage and diversified land use types, thus improving the generalization ability of the model. In addition, the construction process of dataset also takes into account the influence of time change and seasonality, and images of some areas have been collected many times in different seasons to ensure that the dataset can reflect the land use changes at different time points. This enhancement of time dimension makes the dataset not only suitable for static land use classification, but also able to support dynamic land use change monitoring and analysis. Generally speaking, the visual language dataset provides a powerful and comprehensive foundation for the semantic segmentation task driven by deep learning by combining optical and SAR images, fine pixel-level labeling, diverse geographical coverage and enhancement of time dimension. Whether used in academic research or practical application, this data set can significantly improve the performance and robustness of the model. The SWJTU-Vision-Language dataset, alongside its corresponding annotations, will be accessible at https://github.com/yeyuanxin110/SWJTU-Vision-Language.

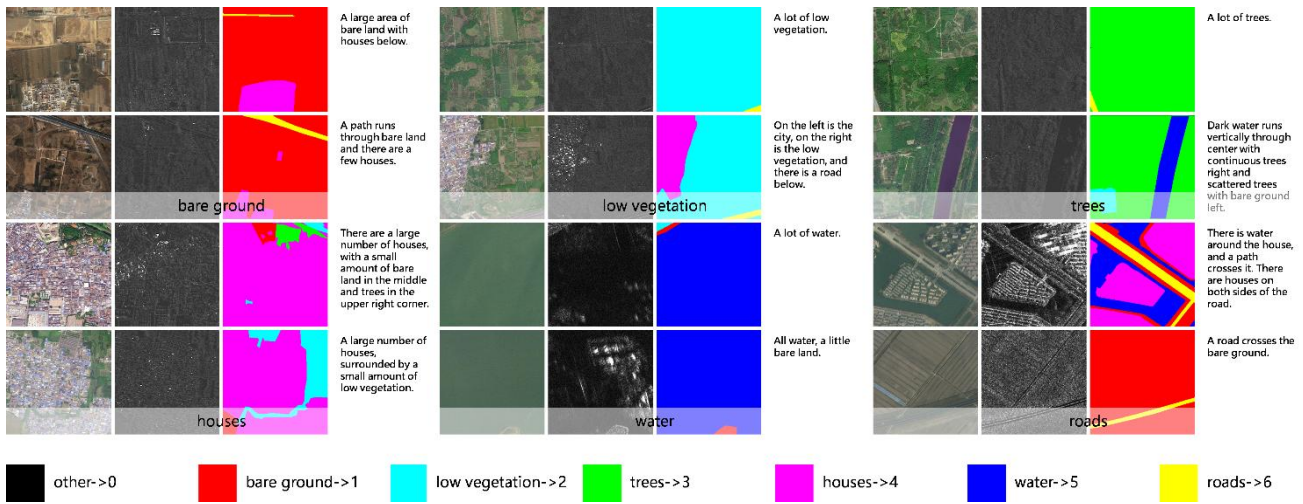

Fig. 4. Representative samples selected from SWJTU-Vision-Language dataset.

2) YESeg-OPT-SAR dataset: This dataset boasts a 0.5 m spatial resolution and integrates two types of remote sensing imagery: RGB images and SAR images. It comprises 2231 pairs of co-registered $256\times256$-pixel images (covering the same areas) across two distinct study regions. With eight annotated categories, its detailed pixel-level labeling supports precise analysis for diverse applications. We have meticulously annotated each set of images in this dataset with detailed textual descriptions, establishing a reliable benchmark for evaluating the accuracy of our proposed model. Access the dataset on GitHub: https://github.com/yeyuanxin110/YESeg-OPT-SAR.

### 4.2 Implementation Details

All our experiments were smoothly conducted using a NVIDIA GeForce RTX 3090 GPU and a Intel Core i5-13600KF CPU. This combination provided a solid foundation for our computational needs. To evaluate the performance of our proposed TSMNet, we compare TSMNet with other typical semantic segmentation models (namely, Deeplab V3+, CMGFNet, MCANet, DDHRNet and MSSNet) and the latest DenseCLIP (Rao et al., 2022) and TACOSS (Valérie et al., 2025) model. For these two data sets, we randomly select 800 to train the data, and the remaining data are used as test samples in our experiment. Table I provides an overview of our training and test datasets. To maintain consistency and ensure the accuracy of our findings, we employed the Adam optimizer throughout the experiment. We started with an initial learning rate of 0.1 and completed 100 epochs using a cosine decay schedule. We use 800 pairs of data sets as training data sets, and the rest are at the test level, and the evaluation is carried out at the test level. In order to prevent a small number of "background" categories from affecting the experimental results, we classify them as "other" categories for unified training.

### 4.3 Evaluation Metrics

To assess our proposed framework, we apply overall accuracy ($OA$), $Kappa$ coefficients, mean intersection over union ($mIoU$), $F1$ score, precsion, and recall to gauge performance on our test datasets for various tasks. Here's a quick rundown of the metrics:

$$OA = \frac{\sum_{k=1}^{N} TP_k + TN_k}{\sum_{k=1}^{N} TP_k + FP_k + TN_k + FN_k}, \quad (12)$$

$$Kappa = \frac{OA - p_e}{1 - p_e}, \quad (13)$$

$$mIoU = \frac{1}{N} \sum_{k=1}^{N} \frac{TP_k}{TP_k + FP_k + FN_k}, \quad (14)$$

$$precision = \frac{1}{N} \sum_{k=1}^{N} \frac{TP_k}{TP_k + FP_k}, \quad (15)$$

where $TP_k$, $FP_k$, $TN_k$ and $FN_k$ represents true positive, false positive, true negative, and false negative, respectively, for the class $k$. $n$ is the total number of pixels, and the pixel-wise accuracy $p_e$ is calculated as $\frac{a_1 \times b_1 + ... + a_m \times b_m}{n \times n}$, where $a_m$ and $b_m$ represent the actual number of pixels and the predicted number of pixels of class $m$, respectively.

### 4.4 Accuracy Analysis

We have put eight top-notch semantic segmentation models to the test, sorting them into three neat categories. The first category focuses on models that handle single-modality remote sensing images, like the popular Deeplab V3+. The second group includes multimodal remote sensing image semantic segmentation models, such as CMGFNnet, DDHRNet, MCAnet and MSSNet. The third group is multimodal models of text supervision, such as DenseCLIP and TACOSS.

**4.4.1 Performance verification on SWJTU-Vision-Language dataset**

Table 2: OA and mIoU (%) on the SWJTU-Vision-Language dataset.

| Method | OA | mIoU | User's Accuracy ||||||
|---|---|---|---|---|---|---|---|---|---|
| | | | bare ground | low vegetation | trees | houses | water | roads | others |
| Deeplab v3+ | 65.34 | 37.12 | 63.76 | 52.68 | 51.28 | 72.75 | 80.84 | 58.07 | 38.04 |
| CMGFNet | 64.93 | 34.7 | 63.98 | 59.23 | 50.78 | 76.24 | 74.32 | 66.76 | 32.28 |
| DDHRNet | 62.63 | 32.66 | 62.83 | 54.65 | 50.93 | 79.59 | 81.42 | 56.33 | 33.71 |
| MCANet | 61.42 | 33.5 | 61.03 | 60.02 | 38.31 | 73.06 | 65.13 | 33.82 | 39.48 |
| MSSNet | 65.68 | 37.49 | 74.12 | 58.66 | 32.95 | 78.17 | 62.9 | 25.87 | 12.52 |
| DenseCLIP | 68.09 | 47.44 | **77.17** | 57.47 | 56.8 | 79.53 | 83.11 | 69.91 | 29.65 |
| TACOSS | 65.43 | 42.64 | 74.97 | 58.82 | 39.28 | 73.32 | 65.56 | 44.86 | 20.56 |
| TSMNet | **69.73** | **48.83** | 76.48 | **61.18** | **55.2** | **80.09** | **84.82** | **69.93** | **39.49** |

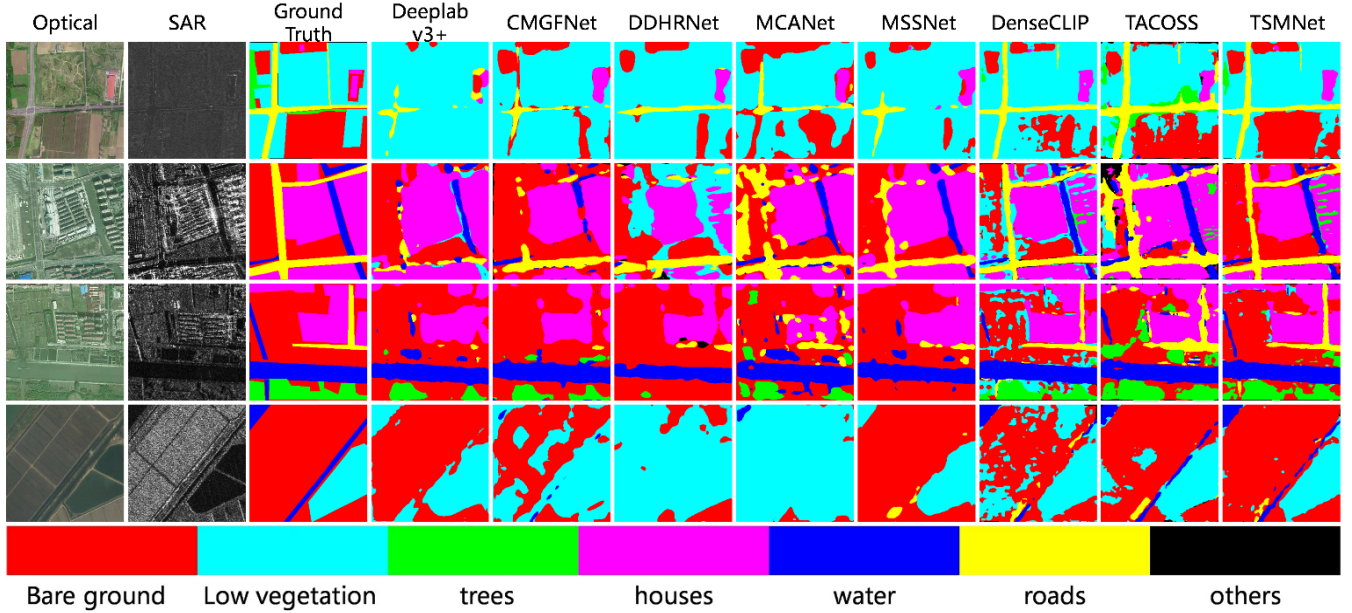

Fig. 5. Visualizations of TSMNet and other networks on the SWJTU-Vision-Language dataset.

To evaluate the performance of TSMNet in multimodal semantic segmentation of open vocabulary, a self-built SWJTU-

Vision-Language dataset is used for experimental verification. As shown in Table 2, TSMNet achieved the highest mIoU of 48.83% under the PyTorch framework, which was 1.39 percentage points higher than the second-best model DenseCLIP. At the same time, the OA of TSMNet reaches 69.73%, which is 1.64% higher than that of the suboptimal model. TSMNet maintains the highest user accuracy in all categories except bare land, especially in the categories of trees (55.2%) and houses (80.09%). TSMNet demonstrated particular success in the challenging road category, where it achieved the top performance. As shown by the qualitative analysis results in Fig. 5, the superior performance of our multi-modal network over its single-modal counterparts directly validates the effectiveness of fusing optical and SAR features for semantic segmentation. However, the study also found that only relying on the fusion of multi-modal image features still has the limitation of insufficient understanding of the real meaning of categories. This, in turn, constrains the model's overall performance on multi-modal segmentation datasets.

Notably, the multi-modal semantic segmentation model based on text and image is significantly better than the semantic segmentation model using multimodal images only. Especially in mIoU index, TSMNet model surpasses DenseCLIP and TACOSS, which highlights the advantages of TSMNet in effectively using the image text description information to significantly improve the accuracy of semantic segmentation.

**4.4.2 Performance analysis on YeSeg-OPT-SAR dataset**

Table 3: OA and mIoU (%) on the YeSeg-OPT-SAR dataset.

| Method | OA | mIoU | User's Accuracy | | | | | | |
|---|---|---|---|---|---|---|---|---|---|
| | | | bare ground | low vegetation | trees | houses | water | roads | others |
| Deeplab v3+ | 78.42 | 51.89 | 82.07 | 79.32 | 58.6 | 69.92 | 92.37 | 64.92 | 42.52 |
| CMGFNet | 78.95 | 53.69 | 83.45 | 73.16 | 66.37 | 76.28 | 76.53 | 67.24 | 47.9 |
| DDHRNet | 79.05 | 53.22 | 81.92 | 68.94 | 35.33 | 67.76 | 87.73 | 38.33 | 46.12 |
| MCANet | 76.91 | 50.9 | 88.67 | 65.8 | 29.31 | 81.58 | 80.61 | 42.72 | 54.35 |
| MSSNet | 80.23 | 55.94 | 89.18 | 73.62 | 46.87 | 79.97 | 82.87 | 44.93 | 59.87 |
| DenseCLIP | 85.13 | 65.47 | 88.47 | 82.53 | 69.06 | **83.43** | 96.04 | 78.58 | 65.96 |
| TACOSS | 77.66 | 56.08 | 73.33 | 57.67 | 39.84 | 65.59 | 77.53 | 44.85 | 40.05 |
| TSMNet | **85.33** | **66.69** | **89.85** | **84.06** | **69.56** | 81.97 | **96.69** | **80.89** | **69.38** |

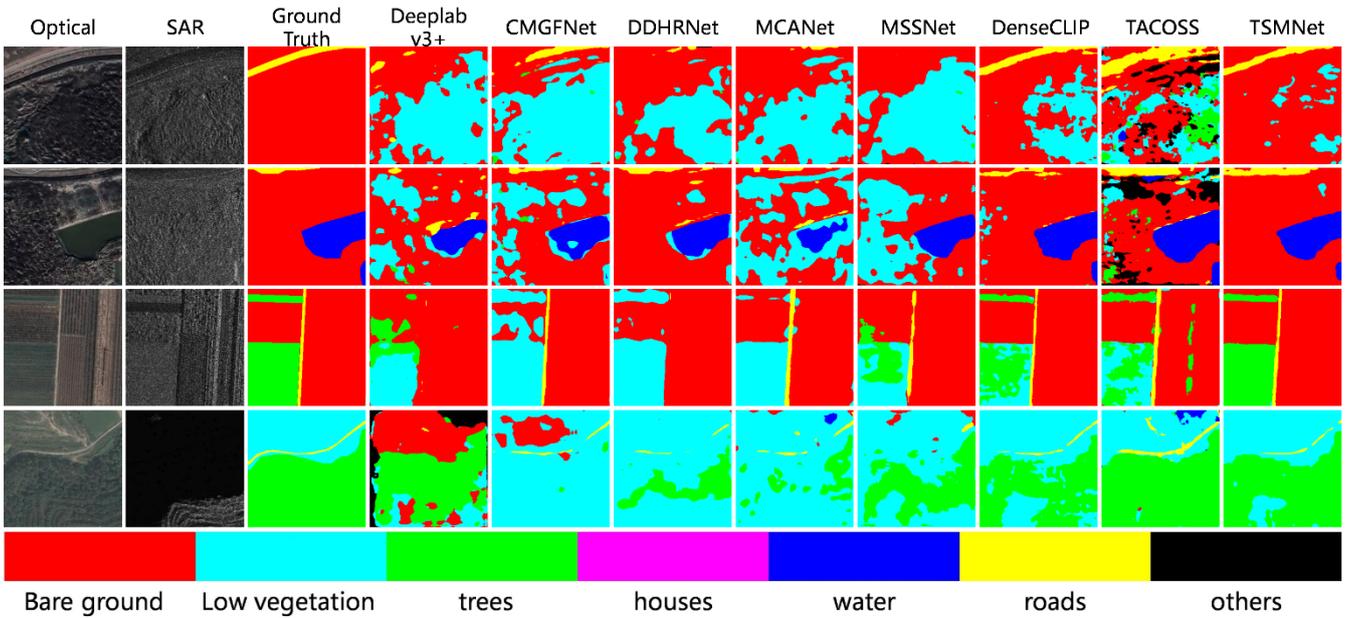

Fig. 6. Visualizations of TSMNet and other networks on the YESeg-OPT-SAR dataset.

In the experiment of YeSeg-OPT-SAR dataset, the models specially designed for remote sensing images, such as Deeplab v3+, CMGFNet, DDHRNet, MCANet, MSSNet, show obvious lack of adaptability when dealing with complex data sets. In contrast, TSMNet achieved the highest mIoU of 66.69% with its multi-modal processing ability, which was 1.22 percentage points higher than the sub-optimal model. In addition, TSMNet also achieved the highest OA of 85.33%. As shown in Table 3, TSMNet achieved the best results in all categories except houses. Compared with the multi-modal image semantic segmentation model, the model guided by natural language shows superior performance, which fully proves the important role of natural language in helping the model understand and distinguish the specific meanings of different categories.

As shown in Fig. 6, when identifying bare land and trees, multimodal image semantic segmentation models tend to misjudge them as low vegetation, which shows that these models only rely on features for classification and fail to fully understand the true meaning of ground objects. In contrast, DenseCLIP, TACOSS and TSMNet, which integrate natural language and remote sensing images, can understand the meaning of categories more accurately, thus achieving more accurate semantic segmentation of open words.

**4.5 Ablation Study**

**4.5.1 Effectiveness of Multi-modal Image Feature Fusion Network**

Table 4: Effects of multi-modal image feature fusion network on accuracy (%).

| Dataset | Network | OA | mIoU |
|---|---|---|---|
| SWJTU-Vision-Language | No | 68.34 | 46.98 |
|  | Yes | **69.73** | **48.83** |
| YESeg-OPT-SAR | No | 83.12 | 65.48 |
|  | Yes | **85.33** | **66.69** |

Section 3 introduces a multi-modal image fusion network designed to integrate optical and SAR images at the feature level. The module begins by aligning multi-scale features from both modalities, followed by feature fusion via an attention mechanism. To evaluate the performance of the fused network, we conduct ablation experiments consisting of two parts, with detailed results provided in Table 4.

As shown in Table 4, incorporating the multi-modal image fusion network substantially improves semantic segmentation accuracy. This approach effectively integrates multi-modal remote sensing data while reducing the potential negative influence of noisy information. The model demonstrates a strong capability to handle complex cross-modal interactions and consistently achieves high segmentation accuracy across diverse multimodal datasets. These results underscore its advantages in mitigating noise, suppressing interference, and resolving semantic inconsistencies during the multimodal fusion process.

### 4.5.2 Effectiveness of Object-level Label Feature Fusion of Image and Text Module and Scene-level Semantic Feature Fusion of Image and Text Module

Table 5: Effects of natural language on accuracy (%). Structure A has no natural language guidance, B has only category features, C has only text description features, and D contains two categories. Bold values indicate the highest accuracy.

| Structure | SWJTU-Vision-Language | | YESeg-OPT-SAR | |
|---|---|---|---|---|
|  | OA | mIoU | OA | mIoU |
| A | 68.31 | 47.99 | 84.97 | 64.79 |
| B | 69.21 | 48.36 | 85.23 | 65.39 |
| C | 68.78 | 48.53 | 85.17 | 65.54 |
| D | **69.73** | **48.83** | **85.33** | **66.69** |

The Section 3 introduces the fusion module of object-level label feature and scene-level semantic feature of image and text features. Natural language makes a significant contribution to dealing with multimodal semantic segmentation. In this paper, two natural languages, category and text description, are aligned and fused with remote sensing images respectively, and the performance of the model is significantly improved. Table 5 presents the detailed experimental results. We can see from the table that the addition of natural language, whether it is category or text description, can significantly improve semantic segmentation performance. Category labels and text descriptions offer complementary alignments at the object-level label and scene-level semantic features, which together help bridge the gap between remote sensing imagery and a realistic, semantic understanding of the scene. The combination of category and text description can further improve the performance of the model. Therefore, TSMNet adopts the method of combining and fusing remote sensing images with natural language pixel-level and global alignment.

## 5. CONCLUSION

In this study, we innovatively combine natural language processing with remote sensing image analysis, and propose a new semantic segmentation method of open vocabulary. Unlike the existing visual language model, which mainly focuses on pixel-level category alignment, our proposed TSMNet framework is more in line with human cognitive laws and emphasizes the process of image understanding from the global to the local. By aligning remote sensing image features, local category features and global description features into the semantic space, the framework realizes hierarchical visual representation learning of multi-level and high-level semantics, and effectively integrates image and text features by using cross-modal attention mechanism. In order to improve the quality of multi-modal feature fusion, we specially designed a multi-modal image feature fusion network, which can not only extract multi-modal features, but also effectively correct feature noise, thus fully retaining the detailed information of the image.

In the aspect of data set construction, we innovatively developed two multi-modal semantic segmentation data set for practical application scenarios, including multi-source data such as optical images, SAR images and text descriptions. At the same time, on the basis of an existing multimodal semantic segmentation data set lacking text information, we have made manual text annotation, which provides more semantic information for model training. The experimental results show that TSMNet has excellent performance on both test data sets. Through visual analysis, the model shows significant advantages in adaptive fusion of multimodal information, which fully verifies the effectiveness of the proposed method. This study not only promotes the development of semantic segmentation technology, but also provides a new research idea for multimodal information fusion, which has important theoretical value and practical application significance.

# 6. ACKNOWLEDGEMENTS

This work was supported by the National Science Fund for Distinguished Young Scholars grant number (No. 62425102)，and are supported by the National Natural Science Foundation of China (No. 42271446)